\def\year{2021}\relax
\documentclass[letterpaper]{article} 
\usepackage{aaai21}  
\usepackage{natbib}  
\usepackage{wrapfig}
\usepackage{lipsum}
\usepackage{subfigure}
\usepackage{times}  
\usepackage{helvet} 
\usepackage{courier}  
\usepackage[hyphens]{url}  
\usepackage{graphicx} 
\usepackage{float}
\usepackage{amsmath}
\usepackage{amsfonts}
\usepackage{color}
\usepackage{comment}
\usepackage{cancel}

\newcommand{\model}{E-BERT}
\newcommand{\ahm}{Adaptive Hybrid Masking}
\newcommand{\sahm}{AHM}
\newcommand{\np}{Neighbor Product Reconstruction}
\newcommand{\snp}{NPR}

\newcommand{\pag}{Product Association Graph}

\newcommand{\ecommerce}{E-commerce}
\newcommand{\tb}{\textbf}
\newcommand{\ti}{\textit}
\newcommand{\mc}{\mathcal}

\newcommand{\bs}{\boldsymbol}
\newcommand{\mt}{\mathtt}

\usepackage[e]{esvect}
\usepackage{booktabs}

\usepackage{array}
\usepackage{bm}
\usepackage{multirow}
\usepackage{caption}
\usepackage[linesnumbered,commentsnumbered,ruled,vlined]{algorithm2e}
\definecolor{darkblue}{rgb}{0.0, 0.0, 0.55}
\definecolor{darkred}{rgb}{0.55, 0.0, 0.0}
\nocopyright

\title{E-BERT: Adapting BERT to \ecommerce~with Adaptive Hybrid Masking\\ and \np}

\author{ Denghui Zhang\textsuperscript{\rm 1}, Zixuan Yuan\textsuperscript{\rm 1}, Yanchi Liu\textsuperscript{\rm 2}, Fuzhen Zhuang\textsuperscript{\rm 3}, \\Haifeng Chen\textsuperscript{\rm 2}, Hui Xiong\textsuperscript{\rm 1}\\}
\affiliations{
 \textsuperscript{\rm 1}Rutgers University, USA, \{denghui.zhang, zy101, hxiong\}@rutgers.edu\\ 
 \textsuperscript{\rm 2}NEC Laboratories America, Inc., USA, {yanchi}@nec-labs.com\\
 \textsuperscript{\rm 3}Institute of Computing Technology, Chinese Academy of Sciences, China\\
}
\begin{document}

\maketitle

\begin{abstract}
Pre-trained language models such as BERT have achieved great success in a broad range of natural language processing tasks.
However, BERT cannot well support E-commerce related tasks due to the lack of two levels of domain knowledge, i.e., phrase-level and product-level. 
On one hand, many E-commerce tasks require accurate understanding of domain phrases, whereas such fine-grained \ti{phrase-level} knowledge is not explicitly modeled by BERT's training objective. 
On the other hand, \ti{product-level} knowledge like product associations can enhance the language modeling of E-commerce, 
but they are not factual knowledge thus using them indiscriminately may introduce noise.
To tackle the problem, we propose a unified pre-training framework, namely, \model. 
Specifically, to preserve phrase-level knowledge, we introduce Adaptive Hybrid Masking, which allows the model to adaptively switch from learning preliminary word knowledge to learning complex phrases, based on the fitting progress of two modes.
To utilize product-level knowledge, we introduce \np, which trains \model~to predict a product's associated neighbors with a denoising cross attention layer.
Our investigation reveals promising results in four downstream tasks, i.e., review-based question answering, aspect extraction, aspect sentiment classification, and product classification.
\end{abstract}

\section{Introduction}
Unsupervised pre-trained language models like BERT \cite{devlin2019bert} have greatly advanced the natural language processing research in recent years.
However, these models are pre-trained on open-domain corpus and then fine-tuned for generic tasks, 
thus cannot well support domain-specific tasks.  
To this end, several {domain-adaptive} BERTs have been proposed recently, such as BioBERT \cite{lee2020biobert}, SciBERT \cite{beltagy2019scibert}.
They employ large-scale domain corpora to obtain language knowledge of specific domains,
e.g., BioBERT uses 1M PubMed articles for pre-training.
\begin{figure}[t]
\setlength{\belowcaptionskip}{-6pt}
\centering
\includegraphics[width=0.47\textwidth]{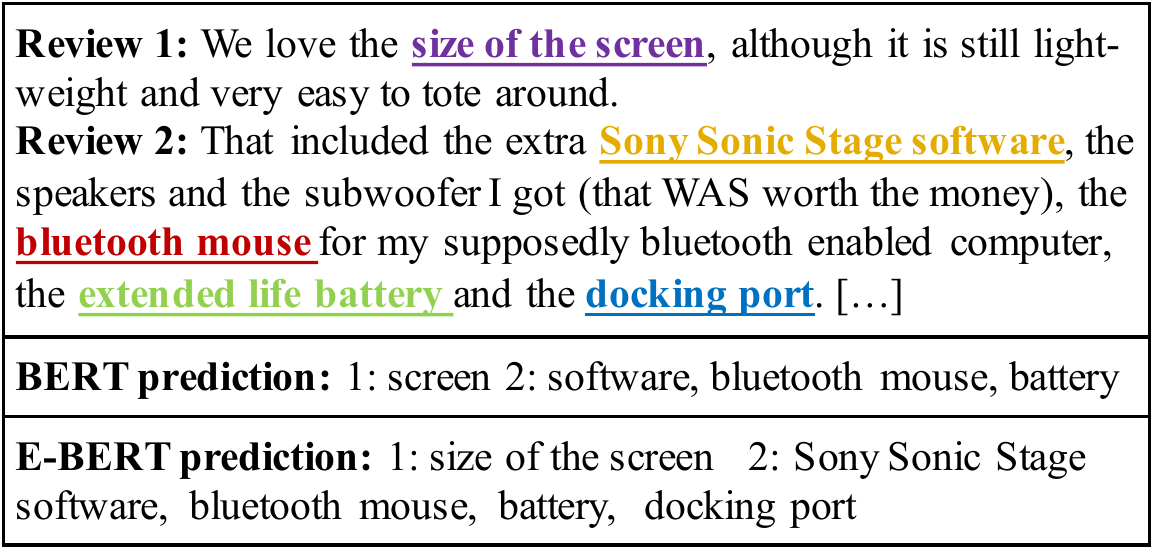}
\vspace{-0.4cm}
\caption{An example of review aspect extraction, with answers marked in color. The vanilla BERT tends to make incomplete/wrong predictions, and miss aspects sometimes. But it gets improved after integrating phrase-level and product-level domain knowledge.}
\label{intro}
\vspace{-0.3cm}
\end{figure}
Despite this, 
they adopt the same architecture and pre-training approach as BERT does, 
neglecting the crucial domain knowledge which is beneficial for downstream tasks.
Specifically, we find two levels of domain knowledge may not be effectively captured by BERT, i.e., \ti{phrase-level} and \ti{product-level} knowledge.
Along this line, we incorporate these two-levels of domain knowledge to BERT for the E-commerce domain and perform evaluations on several related tasks.

First, many tasks in E-commerce involve understanding various \ti{domain phrases}.
However, such fine-grained \ti{phrase-level} knowledge is not explicitly captured by BERT's training objective, i.e., Masked Language Model (MLM).
Specifically, MLM aims to predict individual masked words from incomplete input, 
thus being a \ti{word-oriented} rather than \ti{phrase-oriented} task.
Although some subsequent work \cite{sun2020ernie} proposes to mask phrases so that BERT can learn phrase-level knowledge, there are two major limitations: 
(\romannumeral1) they mask phrases that are simply obtained by chunking, may not be domain-specific;
(\romannumeral2) they discard word masking after using phrase masking, yet, we argue word-level knowledge is preliminary to phrase understanding.
Figure \ref{intro} gives a motivating example from review {A}spect {E}xtraction. 
The task aims to extract entity aspects on which opinions have been expressed.
Without phrase modeling, BERT tends to miss aspects or output incomplete aspects.
It gets improved after effective phrase knowledge encoding.

\begin{figure}[t]
\setlength{\belowcaptionskip}{-6pt}
\centering
\includegraphics[width=0.465\textwidth]{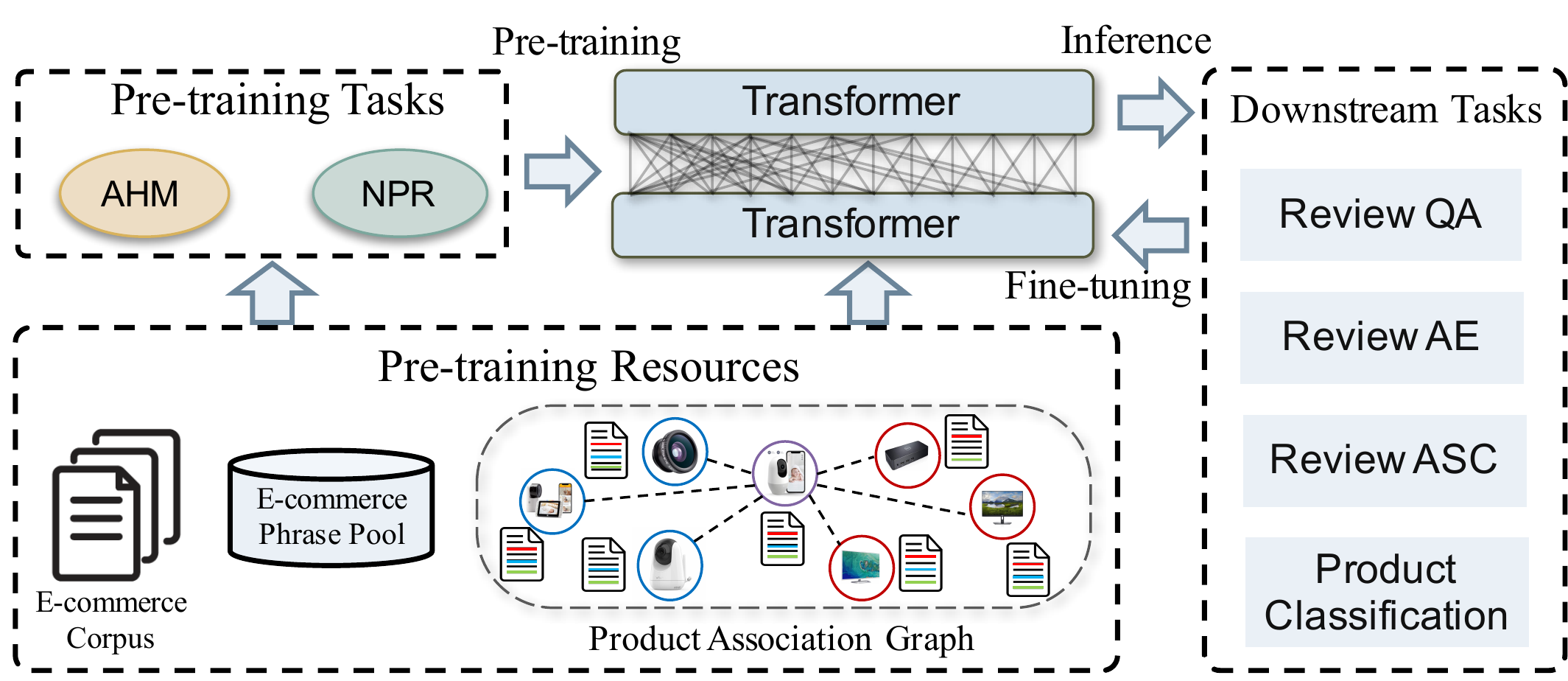}
\vspace{-0.2cm}
\caption{Overview of \model.
}
\label{frame}
\vspace{-0.1cm}
\end{figure}
\begin{figure}[t]
\setlength{\belowcaptionskip}{-6pt}
\centering
\includegraphics[width=0.46\textwidth]{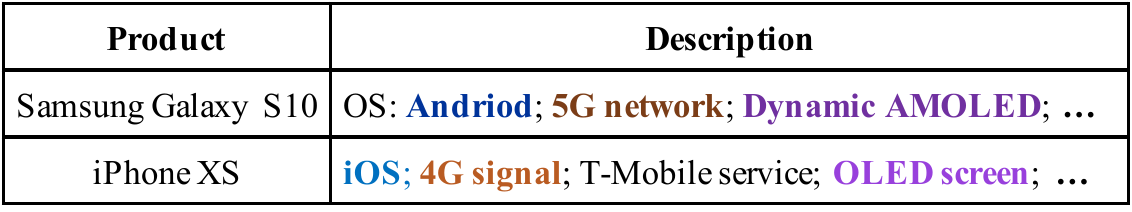}
\vspace{-0.2cm}
\caption{Two associated products and their descriptions.
\vspace{-0.3cm}
}
\end{figure}
On the other hand, there is rich semantic knowledge hidden in product associations, and we consider it as \ti{product-level} knowledge.
Existing models like BERT rely on co-occurrence to capture the semantics of words and phrases, which is  inefficient and expensive.
For instance, to teach the model that \texttt{Android} and \texttt{iOS} are semantically similar and correlated, a large number of co-occurrences of them are required in the corpus.
Leveraging product associations to bridge the contents of \texttt{Samsung galaxy} and \texttt{iPhone}, we can easily enhance such semantic learning. 
However, it is challenging in practice because different fragments in two connected contents have different association confidence, without differentiating, it may introduce extra noise. 

To enable pre-trained language models with the above two levels of domain knowledge, 
we propose a unified pre-training framework, \model.
As shown in Figure \ref{frame}, we continue to use Transformer as the underlying architecture,
and leverage a massive domain corpus, a high-quality E-commerce phrase pool, and a product association graph as our pre-training resources.
To train \model~on them,
we introduce two novel improvements as the pre-training tasks, i.e., Adaptive Hybrid Masking (\sahm) and \np~(\snp):

(1) \sahm~extends MLM by introducing a new masking strategy.
Specifically, it sets two different modes, i.e., word masking mode and phrase masking mode.
The former randomly masks separate words while the latter masks domain phrases.
Moreover, it can adaptively switch between the two modes based on feedback losses,
enabling the model to capture word-level and phrase-level knowledge progressively.

(2) In \snp, we train \model~to reconstruct the neighbor products in the association graph given a central product, using its own content representation and a de-noising cross attention layer. 
The cross attention layer enables the model to pay more/less attention to different positions of the content according to their relevance.
As a result, NPR transforms product-level knowledge into semantic knowledge without introducing too much noise.

To validate the effectiveness of the proposed approach, we fine-tune \model~on four downstream tasks, i.e., Review-based Question Answering (RQA), Aspect Extraction (AE), Aspect Sentiment Classification (ASC), and Product Classification.
The experimental results show that \model~significantly outperforms BERT and several following work on these domain-specific tasks, by taking full advantage of the phrase-level and product-level knowledge. 


\section{Methodology}
\begin{table}[t]
\renewcommand\arraystretch{1}
\small
  \centering
\caption{High-quality phrases of 6 product categories.}
\vspace{-0.1cm}
  \centering
  \begin{tabular}{|m{2.3cm}<{\centering}|m{5cm}|}
    \hline
    \tb{Category} & \tb{\quad \quad \quad Top-rated phrases} \\
    \hline
    \hline
     Automotive& jumper cables, cometic gasket, angel eyes, drink holder, static cling  \\
    \hline
     Clothing, Shoes and Jewelry&  high waisted jean, nike classic, removable tie, elegant victorian, vintage grey\\
     \hline
    Electronics & ipads tablets, SDHC memory card, memory bandwidth, auto switching \\
    \hline
    Office Products & decorative paper, heavy duty rubber, mailing labels, hybrid notebinder \\
    \hline
    Sports and Outdoors & basketball backboard, table tennis paddle, string oscillation, fishing tackles \\
    \hline
    Toys and Games & hulk hogan, augmented reality, teacup piggies, beam sabers, naruto uzumaki \\
    \hline
  \end{tabular}
\label{phrase}
\vspace{-0.2cm}
\end{table}
In this section, we first present the pre-training resources used in \model.
Then, we provide an in-depth introduction about our improvements in pre-training, i.e., \ahm~and \np.

\subsection{Pre-training Resources}
\subsubsection{E-commerce Corpus}
We extract millions of product titles, descriptions, and reviews from the Amazon Dataset\footnote{https://nijianmo.github.io/amazon/index.html} \cite{ni2019justifying} to build this corpus.
We divide the corpus into two sub-corpus, i.e., product corpus and review corpus.
In the first corpus, each line corresponds to a product title and its description, while in the second, it corresponds to a user comment on a specific product.
The corpus serves as the foundation for \model~to learn preliminary language knowledge.

\begin{figure*}[t]
\setlength{\belowcaptionskip}{-6pt}
\centering
\includegraphics[width=0.95\textwidth]{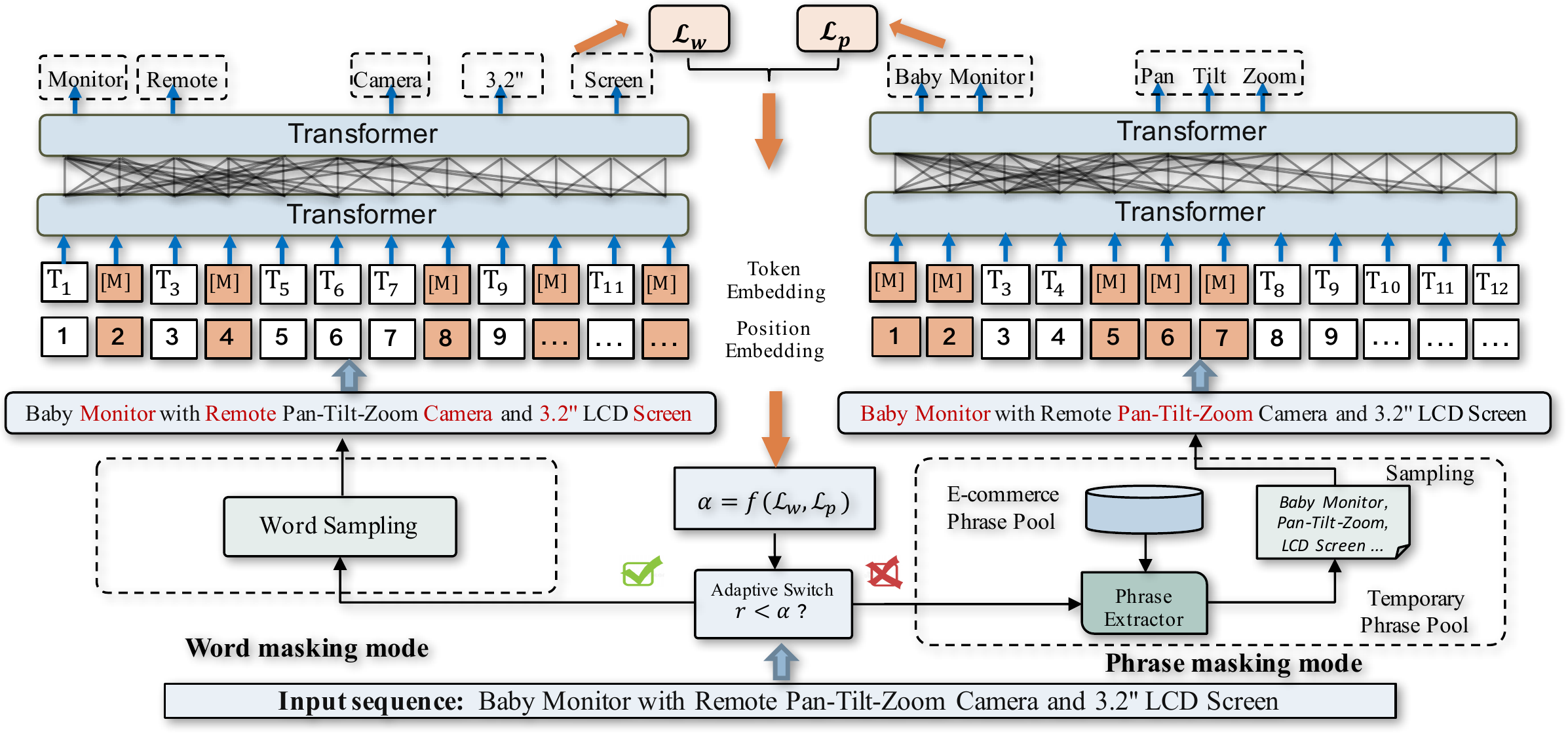}
\vspace{-0.2cm}
\caption{
The illustration of \ahm. 
Based on the feedback losses, it adaptively switches between two masking modes,
enabling the model to learn word-level and phrase-level knowledge in a progressive manner.
}
\label{ahm}
\vspace{-0.1cm}
\end{figure*}
\subsubsection{E-commerce Phrase Pool}
To incorporate domain phrase knowledge into \model, we extract plenty of domain phrases from the above corpus and build an E-commerce phrase pool in advance.
Considering phrase quality,
we adopt AutoPhrase\footnote{https://github.com/shangjingbo1226/AutoPhrase}, 
a high efficient phrase mining method \cite{shang2018automated}, 
can generate a quality score for each phrase based on corpus-level statistics like \ti{popularity, concordance, informativeness,} and \ti{completeness}.
In total, we extract more than one million initial phrases.
Then,
we filter out phrases where $\mt{score<0.5}$ to get quality phrases and store them in the pool.
We also attach the score of each phrase in the pool, which is used for phrase sampling in \sahm.
Table \ref{phrase} shows the top-ranked phrases from six product categories.
Compared with existing work using chunking to select noun phrases for masking, our phrase pool is more diversified, fine-grained, and domain-specific.
Discussion about the effects of choosing different phrase sets is shown in the experiment section.

\subsubsection{\pag}
To enable \model~with product-level knowledge, we build Product Association Graph in advance.
Specifically, products in our corpus are represented as nodes, and associations among products are represented as undirect edges.
To obtain product associations, we extract product pairs such as substitutable and complementary from the Amazon dataset, using a heuristic method based on consumer shopping statistics \cite{mcauley2015inferring}.

\subsection{\ahm}
To encode phrase-level knowledge effectively, we introduce a new masking strategy, namely, \ahm~(\sahm), 
which is easy to implement by extending MLM.
In \sahm,
we set two masking modes, i.e., \ti{word masking} and \ti{phrase masking} respectively.
The former masks word units while the latter masks domain phrase units, resulting in inconsistent difficulty levels of reconstructing the masked tokens.
To this end,
we adaptively switch from predicting masked words to predicting masked phrases, enabling \model~to capture word-level and phrase-level knowledge in a progressive manner.

\subsubsection{Word Masking Mode}
In this mode, we select random words from input sequence iteratively until obtain 15\% tokens for masking.
This scheme learns preliminary word-level semantics, which is essential for phrase understanding. 

\subsubsection{Phrase Masking Mode}
In phrase masking mode, we randomly mask consecutive tokens that can form \ti{quality domain phrases}.
Specifically, given an input sequence of tokens $X=\{x_i\}_{i=1}^{n}$, 
we first detect all the E-commerce phrases $\{p_i\}_{i=1}^{m}$ in $X$, leveraging the E-commerce phrase pool $\mc{P}_E$ along with a rule-based phrase matcher\footnote{https://spacy.io/usage/examples\#phrase-matcher}.
Then, we create a temporary phrase pool $\mc{P}_X$ consisting of the detected phrases $\{p_i\}_{i=1}^{m}$.
Some input sequences may contain too few domain phrases,
therefore, to ensure we have enough and diverse phrases, we extend $\mc{P}_X$ with noun phrases.
That is, we extract all the noun phrases $\{n_i\}_{i=1}^{l}$ in $X$ using constituency parsing\footnote{https://spacy.io/universe/project/self-attentive-parser}, 
and abandon the ones that have an intersection with the domain phrases $\{p_i\}_{i=1}^{m}$.
Then, we add the rest ``clean'' noun phrases in to $\mc{P}_X$.
Based on the extended $\mc{P}_X$, we sample phrases iteratively until obtain approximately 15\% tokens for masking. 
The probability of selecting each phrase is set as the softmax of its quality score:
\begin{equation}
p(p_i)=\frac{\exp\big(s[p_i]\big)}{\sum_{p_j\in\mc{P}_X}\exp\big(s[p_j]\big)},
\end{equation}
where $s[p_i]$ denotes the score of phrase $p_i$.
For the supplemental phrases, it is assigned with the lowest score in $\mc{P}_E$.
Phrases with higher scores are usually more E-commerce related, 
thus the quality-based sampling impels our model to pay more attention to unique domain phrases.

\subsubsection{Adaptive Switching}
When learning a new language, people usually start with individual words (the vocabulary), and gradually turn to study more complex phrases and expressions.
Inspired by this, we start pre-training with word masking mode, 
and set a time-varying parameter $\alpha$ to adaptively switch to phrase mode.

In detail, at each iteration ($t^{th}$), we calculate a ``fitting index'' for both modes to track their fitting progress, i.e., $\eta_{w}^{t}$ and $\eta_{p}^{t}$.
The larger $\eta_{w}^{t}$ ($\eta_{p}^{t}$) is, the less sufficient the model is trained on the word (phrase) mode.
Next, we calculate $\gamma^{t}$, representing the relative importance of the word mode, and rescale it to [0,1] via a non-linear unit ($\mathtt{tanh}$) to get the probability of choosing word mode at the next iteration ($t+1^{th}$), i.e., $ \alpha^{t+1}$:
\begin{equation}
\eta_{w}^{t}=\frac{\Delta^{t,t-1}_{w}}{\Delta^{t,1}_{w}}=\frac{\big[\mc{L}_{w}^{t-1}-\mc{L}_{w}^t\big]_{+}}{\mc{L}_{w}^{1}-\mc{L}_{w}^t},
\end{equation}
\begin{equation}
\eta_{p}^{t}=\frac{\Delta^{t,t-1}_{p}}{\Delta^{t,1}_{p}}=\frac{\big[\mc{L}^{t-1}_{p}-\mc{L}^t_{p}\big]_{+}}{\mc{L}^{1}_{p}-\mc{L}^t_{p}},
\end{equation}
\begin{equation}
\gamma^{t}=\frac{\eta_{w}^{t+1}}{\eta_{p}^{t+1}},
\end{equation}
\begin{equation}
 \alpha^{t+1}=\mathtt{tanh}(\gamma^{t}).
\end{equation}
where $\Delta^{t,t-1}_{w}$ denotes the loss reduction of word mode between current and last iteration.  $\Delta^{t,1}_{w}$ denotes the total loss reduction.
$\mc{L}_{w}^t$ denotes the loss and will only be updated if word mode is selected at the $t$-th iteration.
$\Delta^{t,t-1}_{p}$, $\Delta^{t,1}_{p}$, denotes the counterparts in phrase mode.
$[x]_+$ is equivalent to $max(x,0)$.
When $\eta_{w}^{t+1}\gg\eta_{p}^{t+1}$, $ \alpha^{t+1}\approx1$, and the word mode becomes dominating, vice versa.
In other words, $\alpha$ controls the model to switch to the weaker mode adaptively.

Figure \ref{ahm} presents an overall illustration of \sahm. 
At each iteration,
we first generate a random number $r\in[0,1]$, then we select word masking mode if $r<\alpha^t$ and select phrase mode otherwise. 
To be noted, for the first $t\leq T_1$ iterations, we set $\alpha^t=\alpha_0$ ($\alpha_0>0.5$) to make word mode more likely to be selected.
After this initial stage, $\eta_{p}$ gets larger than $\eta_{w}$ and $\alpha$ decreases, consequently, the probability of selecting phrase mode gets larger.
Until the end, it will switch between the two modes adaptively based on their losses, balancing word-level and phrase-level learning.

\subsubsection{Reconstructing Masked Tokens}
For both modes after selecting tokens to be masked,
following BERT to mitigate the mismatch between pre-training and fine-tuning,
we replace the selected tokens with 
(1) the \texttt{[MASK]} token 80\% of the time, (2) a random token 10\% of the time, (3) the original token 10\% of the time. 
Next, we predict each masked token by feeding their output embedding to a shared softmax layer (take word masking mode as example), i.e.,
\begin{equation}
p\big(\footnotesize{ X^{t}_m\big|X^{t}_{\backslash \mc{W}_{X^t}}}\big)=\displaystyle\frac{\exp\left(\bs{W_m^\top}\big[\mathtt{E}\texttt{-}\mathtt{BERT}\big(X^{t}_{\backslash \mc{W}_{X^t}}\big)\big]_m\right)}{\displaystyle\sum\limits_{k\in \mc{V}}\exp\left(\bs{W_k^\top}\big[\mathtt{E}\texttt{-}\mathtt{BERT}\big(X^{t}_{\backslash \mc{W}_{X^t}}\big)\big]_k\right)},
\end{equation}
where $\bs{W}$ represents the parameters of softmax layer.
$\mc{V}$ denotes the vocabulary.
$X^t$ denotes the input sequence.
$\mc{W}_{X^t}$ denotes the set of masked tokens in word masking mode, 
$\backslash$ denotes set minus,
$X^{t}_{\backslash \mc{W}_{X^t}}$ denotes the modified input where $\mc{W}_{X^t}$ are masked.
$X^{t}_m$ denotes the masked token to be predicted, $m\in\mc{W}_{X^t}$.
The overall loss function of \sahm~is the combined cross entropy of the two masking modes, i.e.,
\begin{equation}
\begin{aligned}
\mc{L}_{\text{AHM}} = -\small\frac{1}{|\mc{D}|}&\sum_{X^{t}\in\mc{D}}\alpha^{t}\log\prod_{m\in{\mc{W}_{X^t}}} p\big(X^{t}_m\big|X^{t}_{\backslash \mc{W}_{X^t}}\big)+\\
&(1-\alpha^{t})\log\prod_{m\in{\mc{P}_{X^t}}} p\big(X^{t}_m\big|X^{t}_{\backslash \mc{P}_{X^t}}\big),
\end{aligned}
\end{equation}
where $\mc{D}$ represents the training corpus. 
$p\big(\footnotesize{ X^{t}_m\big|X^{t}_{\backslash \mc{P}_{X^t}}}\big)$ denotes the prediction in phrase masking mode, calculated the same way as word masking mode.
$\mc{P}_{X^t}$ denotes the set of masked tokens in phrase mode.


\begin{figure}[t]
\setlength{\belowcaptionskip}{-6pt}
\centering
\includegraphics[width=0.46\textwidth]{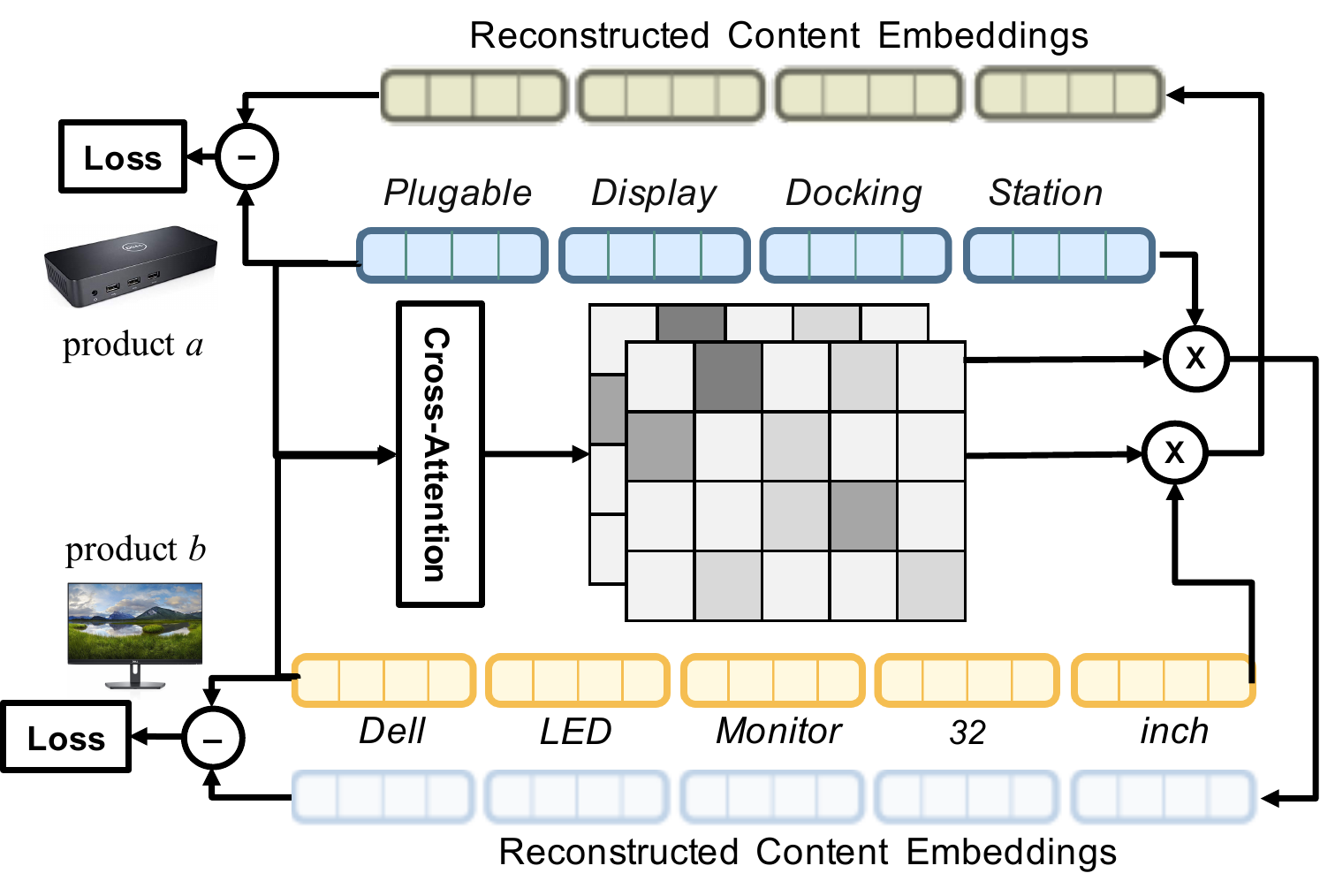}
\vspace{-0.2cm}
\caption{Illustration of \np.}
\label{np}
\vspace{-0.2cm}
\end{figure}
\subsection{\np}
In this task, 
we train E-BERT to reconstruct the neighbor (associated) product's content using the central product's content, and thereby, 
transform the hidden semantic knowledge into the weights of the model.

As illustrated in Figure \ref{np},
we first sample associated products from the product association graph and put their content embeddings into a pair (shown in the the middle of the figure).
The cross attention layer is then used to learn a correlation matrix, indicating word-level correlations between two products.
Next, we multiply the correlation matrix with the central product's content embeddings to generate a set of reconstructed content embeddings for the neighbor product.
They are optimized to resemble the real ones via the content reconstructing loss.
Considering the central product can also be viewed as neighbor to the reconstructed product, we perform the reconstructing task in both directions.
\subsubsection{Content Embeddings}
Given the product pair $(a,b)$, we feed their contents (title and description) into \model~respectively to get their content embeddings, i.e.,
\begin{equation}
\begin{aligned}
\{\bs{w_i}\}_{i=1}^n = \mathtt{E}\texttt{-}\mathtt{BERT}\big(\{\bs{a_i}\}_{i=1}^n\big),\\
\{\bs{o_i}\}_{i=1}^n = \mathtt{E}\texttt{-}\mathtt{BERT}\big(\{\bs{b_i}\}_{i=1}^n\big).
\end{aligned}
\end{equation}
\subsubsection{Cross Attention Layer}
We adopt the cross attention layer to generate two correlation matrices, i.e.,
\begin{equation}
\begin{aligned}
\alpha_{ij} = \frac{\exp{(\bs{w_i}\bs{o_j})}}{\sum_{j'} \exp{(\bs{w_i}\bs{o_{j'}})}}, \;
\beta_{ji} = \frac{\exp{(\bs{w_i}\bs{o_j})}}{\sum_{i'} \exp{(\bs{w_{i'}}\bs{o_{j}})}} ,
\end{aligned}
\end{equation}
where $w_i$ indicates the $i^{th}$ word in the product $a$'s content and $o_j$ represents the $j^{th}$ word in $b$. 
The cross attention weight $\alpha_{ij}$ and $\beta_{ji}$ both indicates the correlation between $w_i$ and $o_j$, but using different normalizers.
\subsubsection{Reconstructed Embeddings}
Using the attention weights, we compute a weighted average of the real content embeddings, i.e.,
\begin{equation} \label{eq:aligned_features}
\begin{aligned}
\bs{w}'_i = \sum_{j} \alpha_{ij} \bs{o}_j \;,\; 
\bs{o}'_j = \sum_{i} \beta_{ji}  \bs{w}_i .
\end{aligned}
\end{equation}
where $\bs{w}'_i$ and $\bs{o}'_j$ are the reconstructed embeddings, they are subsequently optimized to resemble the original content embeddings $\bs{w}_i$ and $\bs{o}_j$. 
The negative effect of noise is minimized by the cross attention as it automatically assigns smaller weights to irrelevant contents.
\subsubsection{Reconstructing Loss}
We define the content reconstructing loss of a product pair as the Euclidean distance between their real content embeddings to the reconstructed embeddings, i.e.,
\begin{equation}
    \begin{aligned}
        \langle a,b \rangle &= \sum_{i} ||\bs{w}_i-\bs{w}'_i||_2^2 + \sum_{j} ||\bs{o}_j-\bs{o}'_j||_2^2 \\
    \end{aligned}
\end{equation}
We use a triplet loss as the final loss to pull relevant product-product pairs close while pushing irrelevant ones apart:
\begin{equation} 
    \begin{aligned}
        \mathcal{L}_{\text{\snp}} &= \max (0, 1+\langle a,b \rangle-\langle a,b^- \rangle), 
    \end{aligned}
\end{equation}
where $b^-$ is a randomly sampled negative product that is not related to $a$. 

To be noted, we only train \snp~on the product corpus where the input is formated as content pairs $\langle\text{content}(a),\text{content}(b)\rangle$.
We do not train \snp~on the review corpus since it consists of user feedbacks, can not reflect product semantics accurately.

\section{Experiments}
In this section, we conduct extensive experiments to answer the following research questions:
\begin{itemize}
	\item What is the performance gain of the E-commerce corpus for each downstream task, with respect to the state-of-the-art performance?
	\item What is the overall performance gain of our pre-training framework incorporating two levels domain knowledge?
	\item What is the performance gain of each component (i.e., \sahm~and \snp) in \model?
\end{itemize}
\subsection{Baselines}
In this paper, we compare \model~to the following baseline methods:
\begin{itemize}
	\item \tb{BERT-Raw:}~The vanilla BERT which is pre-trained on large-scale open-domain corpus\footnote{We use the pre-trained model released by Huggingface.}. We use this baseline to answer the first question.
	\item \tb{BERT:}~The vanilla BERT which is further post-trained on our E-commerce corpus.
	We compare with this baseline to answer the second question.
	\item \tb{BERT-NP:}~The vanilla BERT which is post-trained on our E-commerce corpus, 
	but uses a different masking strategy, i.e., masks noun phrases instead of words. We use this to validate the effect of our domain phrase pool.
	\item \tb{SpanBERT}: An variant of BERT which masks spans of tokens instead of seperate tokens. We compare with it to further validate the effect of the phrase masking scheme.
\end{itemize}
For ablation studies, we further compare with the following internal baselines:
\begin{itemize}
	\item \tb{E-BERT-DP:}~The reduced \model~which only uses the phrase masking mode, without word-level masking.
	\item \tb{E-BERT-\sahm:}~The reduced \model~which adopts \sahm~to adaptively change the masking mode, but not utilizes~\snp~to encode product-level knowledge.
	\item \tb{E-BERT:}~The full \model, utilizing both \sahm~and \snp~to encode two levels of domain knowledge.
\end{itemize}

\subsection{Pre-training Dataset}
The dataset contains four pre-training resources:
\begin{itemize}
\item\tb{Product Corpus}~ It contains a total of $5,436,547$ product titles and descriptions with a size of $1.4$ GB.
\item\tb{Review Corpus}~ It contains a total of $9,636,112$ million product reviews with a size of $2.3$ GB.
\item\tb{E-commerce Phrase Pool}~ It consists of $536,332$ high quality E-commerce phrases.
\item\tb{Product Association Graph}~ It consists of $2,125,352$ products and $3,484,325$ product associations.
\end{itemize}
\subsubsection{Phrase Overlap}
\begin{wrapfigure}{R}{0.2\textwidth}
\centering
\hspace{-1.5\columnsep}
\includegraphics[width=0.24\textwidth]{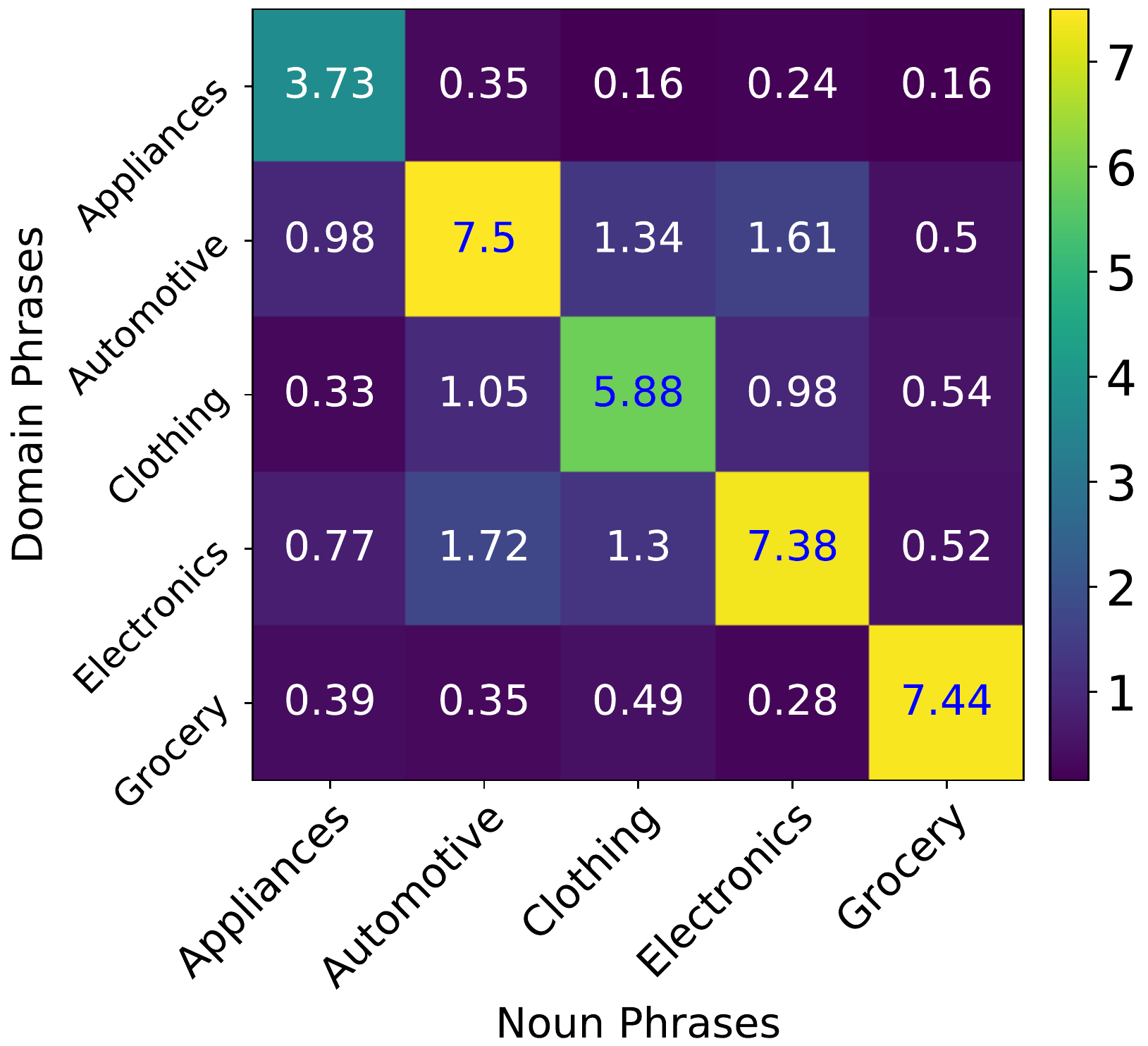}
\vspace{-0.3cm}
\hspace{-1\columnsep}
\caption{The overlap ratio (\%) between the E-commerce phrases and ordinary noun phrases.}
\label{overlap}
\vspace{-0.1cm}
\end{wrapfigure}
Figure \ref{overlap} presents the overlap between the E-commerce domain phrases and the ordinary noun phrases extracted from our corpus, divided by 5 product categories.
Each entry indicates the proportion of domain phrases that also occurred in the noun phrase set.
We can observe that the overlap ratio is low even for the same category, indicating our phrase pool contains more unique phrase-level knowledge.

\begin{table*}[th]
\small
\center
\caption{Results of baselines and our model on E-commerce downstream tasks (\%).}
\label{result}
\begin{tabular}{l|cccc|ccc|ccc|cc}
\toprule
\multicolumn{1}{c}{\multirow{2}{*}{\textbf{Models\textbackslash{}Tasks}}} & \multicolumn{4}{c}{\textbf{Review QA}} & \multicolumn{3}{c}{\textbf{Product Classification}} & \multicolumn{3}{c}{\textbf{Review AE}} & \multicolumn{2}{c}{\textbf{Review ASC}} \\  
\multicolumn{1}{c|}{}                                                         & $P.$            & $R.$           & $F1$     &$EM$      & $Acc.$          & $MiF1$          & $MaF1$          & $P.$           & $R.$           & $F1$           & $Acc.$                    & $MaF1$           \\ \midrule
\multicolumn{13}{c}{Pre-trained on Wikipedia + BookCorpus by Huggingface.}                                    
\\  \midrule
BERT-Raw                                                                    & 58.91 &   62.58     &    60.69     &  40.22        & 66.54       & 81.90       & 78.82      & 83.15        & 84.66        & 83.90   & 86.01             & 62.87                         \\ \midrule
\multicolumn{13}{c}{Further post-trained on E-commerce corpus by us.}\\       

\midrule                                                                                                                                                               
BERT                                                                & 60.28 &  62.25   & 61.25   & 41.23    & 69.12       & 82.38       & 80.66       & 84.33        & 84.09        & 84.81   & 86.40        &  64.96                 \\
BERT-NP                                                            & 61.39 & 64.57          & 62.94        & 43.35 & 70.28       & 81.72       & 81.34   & 85.23        & 85.71        & 86.11       &  85.79        &  63.21                \\
SpanBERT                                                             & 62.52 &  64.77   &    63.63         & 43.94       & 71.59       & 81.51       & 81.50         & 85.67        & 86.22        & 86.23           & 86.76        &  65.13                \\  \midrule
E-BERT-DP                                                                      & 63.76         & 67.02         & 65.77         & 44.63         &  75.07   & 85.84 &  86.28       & 86.80        & 89.47        & 88.11         &  87.84        &  69.02        \\
E-BERT-\sahm                                                              &  65.18 & 68.30  & 66.18          & \textbf{45.56}        & 76.61   & 86.35 &  {87.32}        & \textbf{87.42}        & \textbf{90.55}       & \textbf{88.96}    & \textbf{89.17}  &  \textbf{70.35}                 \\
E-BERT                                                          & \textbf{66.71} & \textbf{70.13}         & \textbf{68.77}        &  45.40 &   \textbf{78.74}   & \textbf{90.37} &  \tb{90.94}   & 87.35         & 89.61         & 88.42  & 88.43         & 69.32                 \\
\bottomrule
\end{tabular}
\end{table*}
\subsection{Pre-training Details}
All the baselines and \model~is initialized with the weights of the pre-trained BERT (the \texttt{bert-base-uncased} version by Huggingface, with 12 layers, 768 hidden dimensions, 12 heads, 110M parameters). 
We post-train all the baselines except BERT-Raw on the E-commerce corpus for $10$ epochs, with batch size 32 and learning rate 1e-5.
For \model, we adopt Continual Multi-task Learning \cite{sun2020ernie} to combine \sahm~and NPR. 
To be specific, we first train \sahm~alone on the entire corpus for $5$ epochs with the same batch size and learning rate. 
Then, we train \sahm~and NPR jointly on the product corpus for another $5$ epochs.
The only hyperparameter in \sahm, $T_1$, is set to be $1$ epoch.
\subsection{Downstream Tasks}
\subsubsection{Review-based Question Answering}
Given a question $q = \{q_i\}_{i=1}^m$ and a related review  $r = \{r_i\}_{i=1}^n$, it aims to find the span $s = \{r_i\}_{i=s}^e$ from $r$ that can answer $q$.
To fine-tune this task, we adopt the standard BERT approach \cite{devlin2019bert} for span-based QA, which maximizes the sum of the log-likelihoods of the correct start and end positions.
\subsubsection{Review Aspect Extraction}
Given a review $r = \{r_i\}_{i=1}^n$, the task aims to find aspects that reviewers have expressed opinions on.
It is typically formalized as a sequence labeling task \cite{xu2019bert}, in which each token is classified as one of $\{B,I,O\}$, and tokens between $B$ and $I$ are considered the correct aspect.
We apply a dense layer and softmax layer on top of each output embedding to fine-tune.

\subsubsection{Review Aspect Sentiment Classification}
Given an aspect $a = \{a_i\}_{i=1}^l$ and the review sentence $r = \{r_i\}_{i=1}^n$ where $a$ extracted from, this task aims to classify the sentiment polarity (positive, negative, or neutral) expressed on aspect $a$.
For fine-tuning, both $a$ and $r$ are input into \model, and we use the \texttt{[CLS]} token along with a dense layer and softmax layer to predict the polarity.
Training loss is the cross entropy on the polarities.

\subsubsection{Product Classification}
Given a product title $x = \{x_i\}_{i=1}^n$, it aims to classify $x$ using a predefined category hierarchy $\mc{H}$.
Each product may belong to multiple categories, thus making it a multi-lable classification problem.
We use the \texttt{[CLS]} token along with a dense layer and softmax layer to perform prediction.

\subsubsection{Evaluation Datasets}
For review QA, we evaluate on a newly released Amazon QA dataset \cite{miller2020effect}, which consists of $8,967$ samples.
We use the laptop dataset of SemEval 2014 Task 4 \cite{pontiki2016semeval} for both review AE and review ASC tasks, which contains $3,845$ review sentences, $3,012$ annotated aspects and the sentiment polarities on them.
For product classification, we create an evaluation dataset by extracting Amazon product metadata, consisting of $10,039$ product titles and $133$ categories.
For all the datasets, we divide them into training/validation/testing set with the ratio of 7:1:2.

\subsubsection{Fine-tuning Details}
In each task, we adopt the standard architecture for each BERT variant.
We choose the learning rate and epochs from \{5e-6, 1e-5, 2e-5, 5e-5\} and \{2,3,4,5\} respectively.
For each task and BERT variant, we pick the best learning rate and number of epochs on the development set and report the corresponding test results.
We found the setting that works best across most tasks and models is 2 or 4 epochs and a learning rate of 2e-5. 
\begin{figure}[th]
\subfigure[Loss] 
{
	\begin{minipage}{0.47\linewidth}
\includegraphics[scale=0.34]{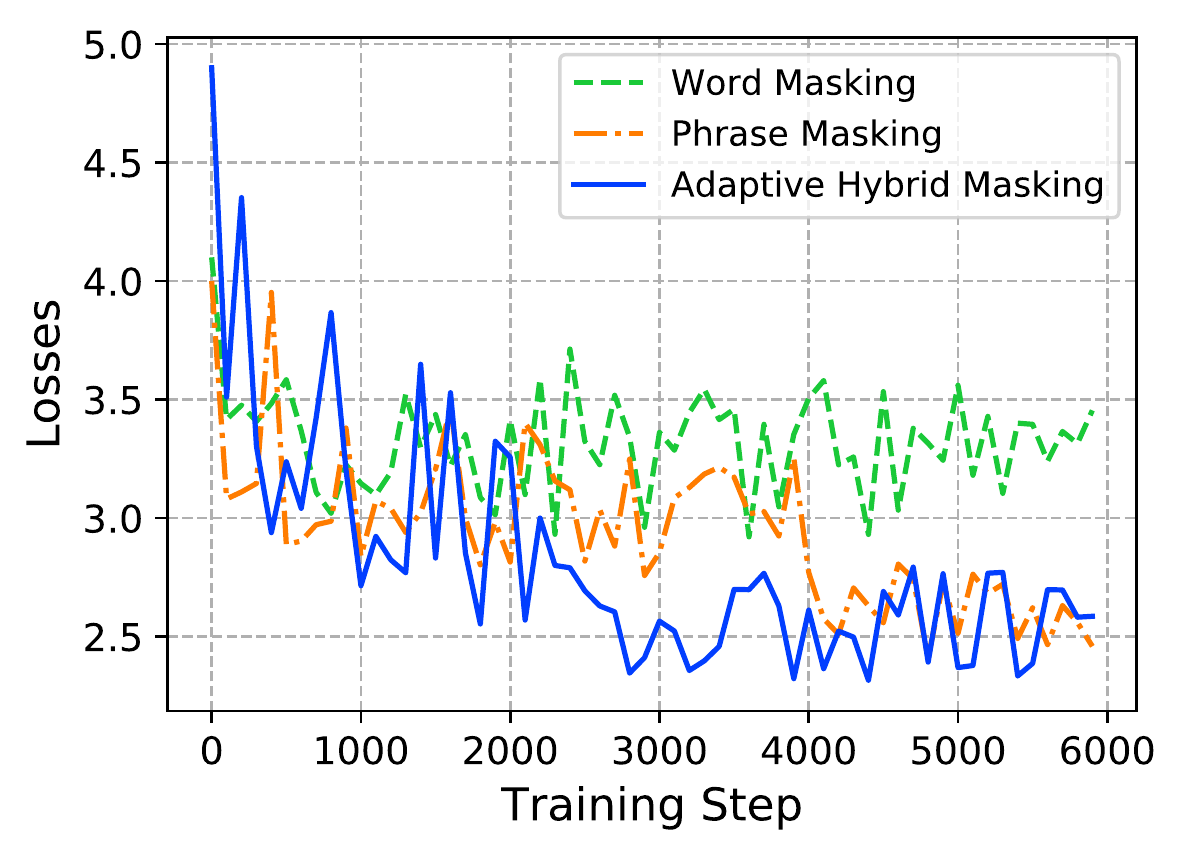}
	\end{minipage}
}
\subfigure[Accuracy] 
{
	\begin{minipage}{0.47\linewidth}
	\centering      
\includegraphics[scale=0.34]{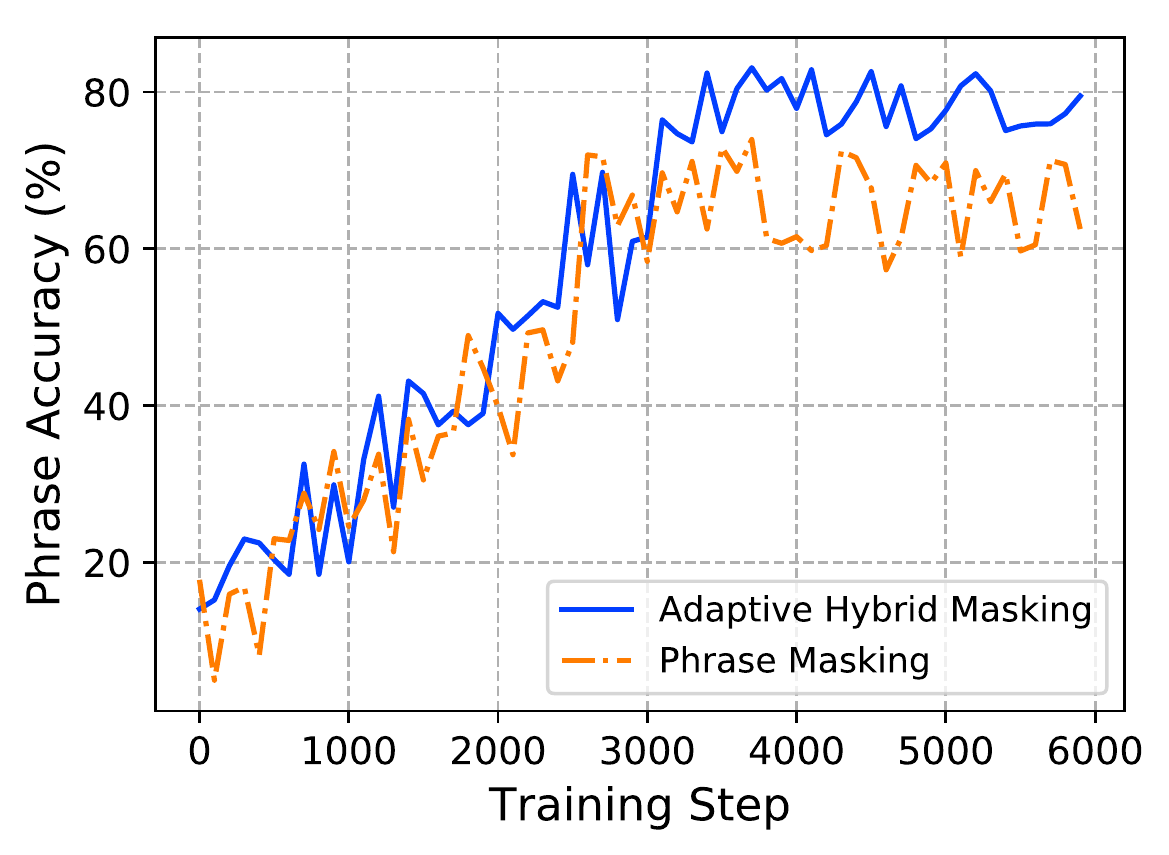}
	\end{minipage}
}
\vspace{-0.2cm}
\caption{The convergence of different masking schemes.} 
\vspace{-0.2cm}
\label{loss}  
\end{figure}
\subsubsection{Evaluation Metrics}
For review QA, we adopt the standard evaluation script from SQuAD 1.1 to report Precision, Recall, F1 scores, and Exact Match (EM).
To evaluate review AE, we report Precision (P), Recall (R), and F1 score.
For review ASC, we report Macro-F1 and Accuracy.
Lastly, we adopt Accuracy (Acc), Micro-F1 (MiF1), and Macro-F1 (MaF1) to evaluate product classification.
\subsection{Result Analysis}
Table \ref{result} presents the results of all the baselines and \model~on the four tasks.
First, we can see that {BERT} outperforms {BERT-Raw} on all the tasks, verifying that the E-commerce corpus can largely improve the performance on related tasks.
Compared with BERT, BERT-NP and SpanBERT achieves further improvements in review QA and review AE, indicating that phrase-level knowledge is quite helpful in these {extractive} tasks.
Comparing \model-DP with BERT-NP and SpanBERT, we prove that our E-commerce phrase pool can provide more quality phrase knowledge for the downstream tasks.
\model~outperforms all the baselines by a large margin in terms of all metrics, verifying the overall superiority of our pre-training framework in E-commerce tasks.
To examine the effectiveness of product-level domain knowledge and each component of \model, we present more discussions in ablation studies.

\subsection{Ablation Studies}
As shown in the bottom of Table \ref{result}, \model-\sahm~consistently outperforms \model-DP in four tasks, 
proving that our adaptive hybrid masking strategy can utilize phrase-level knowledge in a more sufficient way.
Besides, as shown in Figure \ref{loss}, our masking method has a better convergence rate in terms of loss and phrase reconstruction accuracy.
Compared with \model-DP and \model-\sahm, \model~further encodes product-level knowledge via NPR, and it achieves significant improvements in product classification. 
We assume this is because there are strong category correlations between associated products, 
by utilizing product association knowledge, 
\model~enhances feature sharing among different instances.
While the performance of review QA is also boosted, 
review AE and review ASC even deteriorates slightly after using NPR,
indicating product-level knowledge has no significant effect on the task of review aspect analysis.

\subsection{Cross Attention Probing}
\begin{figure}[t]
	\begin{minipage}{0.49\linewidth}
\includegraphics[scale=0.19]{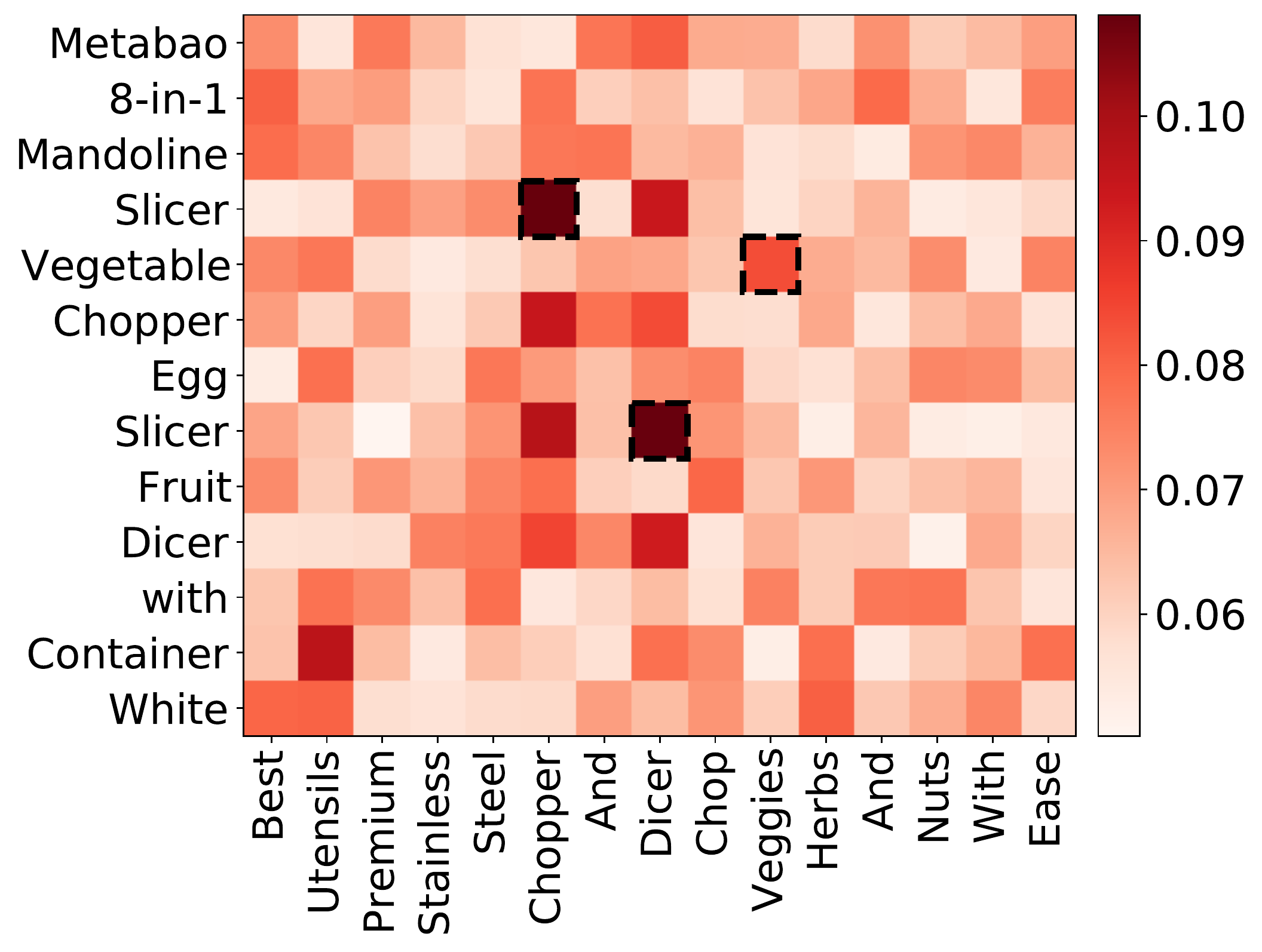}
	\end{minipage}
		\begin{minipage}{0.49\linewidth}
\includegraphics[scale=0.19]{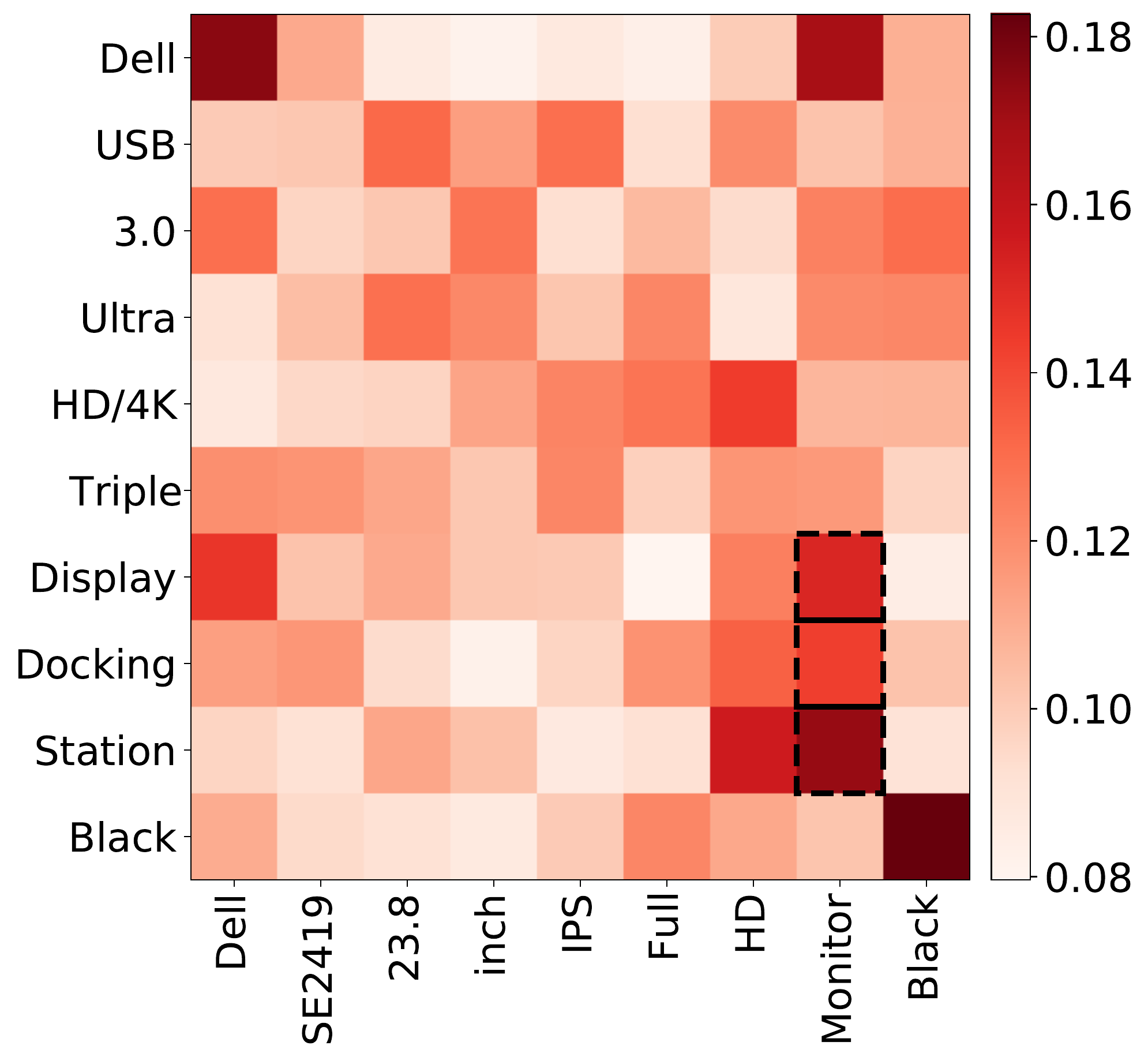}
	\end{minipage}
\vspace{-0.2cm}
\caption{Visualizing the cross attention.} 
\vspace{-0.3cm}
\label{att}  
\end{figure}
Figure \ref{att} presents the cross attention visualization of two pairs of product titles.
In the first example, two similar products \texttt{Mandoline Slicer} and \texttt{Steel Chopper} are connected using the learned attention weights.
The darker color indicates stronger correlations.
It can be seen the cross attention automatically learn to align correlated words in two product contents, e.g., \texttt{(Slicer, Chopper)}, \texttt{(Slicer, Dicer)} and \texttt{(Vegetables, Veggies)}.
In the second example, two complementary products \texttt{Docking station} and \texttt{Dell Monitor} are connected. 
Similarly, correlated contents such as \texttt{(Display, Monitor)} and \texttt{(Docking station, Monitor)} are aligned automatically.


\section{Related Work}
\subsubsection{Pre-trained Language Models}
Recent years have witnessed the great success of Pre-trained Language Models (PLMs) \cite{devlin2019bert,peters2018deep,radford2018improving} on a broad range of NLP tasks.
Compared with traditional word embedding models \cite{gupta2015distributed}, PLMs learn to represent words based on the entire input context to deal with word polysemy, thus captures semantics more accurately.
Following PLMs, many endeavors have been made for further optimization.
SpanBERT \cite{joshi2020spanbert} proposes to reconstruct randomly masked spans instead of single words. 
However, the span consists of random continuous words and may not form phrases, thus fails to capture phrase-level knowledge accurately.
ERNIE-1.0 \cite{sun2019ernie} integrates phrase-level masking and entity-level masking into BERT, which is closely related to our masking strategy.
Unlike them using simple chunking tools to get ordinary phrases, we build a high-quality E-commerce phrase pool and only mask domain phrases.
Besides, we combine word-masking and phrase-masking coherently with Adaptive Hybrid Masking, accelerating the convergence without affecting performance.
Due to space limit, we refer readers to the references for more work along this line \cite{liu2019roberta,lan2019albert,yang2019xlnet,brown2020language,sun2020ernie}.
\subsubsection{Domain-adaptive PLMs}
To adapt PLMs to specific domains, several domain-adaptive BERTs have been proposed recently.
BioBERT \cite{lee2020biobert} and SciBERT \cite{beltagy2019scibert} train BERT on large-scale biomedical and scientific corpus respectively to get a pre-trained languge model for biomedical and scientific NLP tasks.
BERT-PT \cite{xu2019bert} propose to post-train BERT on a review corpus and obtains better performance on the task of review reading comprehension.
\citeauthor{gururangan2020don} propose to continue pre-training on domain corpus as well as task corpus and obtains more performance gains.
More work along this line can be referred to \cite{rietzler2020adapt,ma2019domain,jin2019pubmedqa,huang2019clinicalbert}.
These work only leverages domain corpus for pre-training, without considering special domain knowledge like the product association graph.

\subsubsection{Knowledge Enhanced PLMs}
Recently, to enable PLMs with world knowledge, several attempts \cite{wang2019kepler,peters2019knowledge,zhang2019ernie,liu2020k,wang2020k} have been made to inject knowledge into BERT leveraging Knowledge Graphs (KGs).
Most of these work adopts the ``BERT+entity linking'' paradigm,
whereas, it is not suitable for E-commerce corpus due to the lack of quality entity linkers as well as KGs in this domain.
Instead, we consider utilizing the product association knowledge which is coarse-grained and may introduce noise.
In \model, through Neighbor Product Reconstruction and the de-noising cross attention layer, the meaning of each word in a product content is expanded to those of associated products, greatly enriches the semantic learning.


\section{Conclusions}
In this paper, we proposed a domain-enhanced BERT for E-commerce, namely, \model.
We leveraged two levels of domain knowledge, i.e., phrase-level and product-level, to boost performance on related tasks.
Despite the challenge of modeling phrase knowledge and reducing noise in product knowledge, we provided two technical improvements, i.e., \sahm~and \snp.
Our investigation revealed promising results on four downstream tasks.
Incorporating phrase knowledge via \sahm~can improve the performance significantly on all the investigated tasks.
Utilizing the product-level knowledge via \snp~further boosts the performance on product classification and review QA.

\bibliography{ref}

\end{document}